\documentclass[10pt,twocolumn,letterpaper]{article}

\usepackage{cvpr}              

\usepackage{graphicx}
\usepackage{amsmath}
\usepackage{amssymb}
\usepackage{booktabs}

\usepackage[pagebackref,breaklinks,colorlinks]{hyperref}

\usepackage[capitalize]{cleveref}
\crefname{section}{Sec.}{Secs.}
\Crefname{section}{Section}{Sections}
\Crefname{table}{Table}{Tables}
\crefname{table}{Tab.}{Tabs.}

\def\cvprPaperID{44} 
\def\confName{CVPR}
\def\confYear{2023}

\begin{document}

\title{Real-Time Helmet Violation Detection Using YOLOv5 and Ensemble Learning}

\author{ \thanks{Corresponding and primary author} Geoffery Agorku \\
Department of Civil Engineering\\
University of Arkansas\\
{\tt\small {gagorku}@uark.edu}
\and
 Divine Agbobli \thanks{equal contribution} \\
Department of Civil Engineering\\
Iowa State University\\
{\tt\small {dya}@iastate.edu}
\and
 Vuban Chowdhury \thanks{equal contribution} \& Kwadwo Amankwah-Nkyi \thanks{equal contribution}\\
Department of Civil Engineering\\
University of Arkansas\\
{\tt\small {vubanc,kwadwoa}@uark.edu}
\and
 Adedolapo Ogungbire \thanks{equal contribution} \\
Department of Civil Engineering\\
University of Arkansas\\
{\tt\small {amogungb}@uark.edu}
\and
Portia Ankamah Lartey\\
Transportation Engineer\\
Gorove Slade Associates\\
{\tt\small {plartey}@asu.edu}
\and 
Armstrong Aboah\\
Department of Radiology\\
Northwestern University\\
{\tt\small {armstrong.aboah}@northwestern.edu}
}
\maketitle

\begin{abstract}
   The proper enforcement of motorcycle helmet regulations is crucial for ensuring the safety of motorbike passengers and riders, as roadway cyclists and passengers are not likely to abide by these regulations if no proper enforcement systems are instituted. This paper presents the development and evaluation of a real-time YOLOv5 Deep Learning (DL) model for detecting riders and passengers on motorbikes, identifying whether the detected person is wearing a helmet. We trained the model on 100 videos recorded at 10 fps, each for 20 seconds. Our study demonstrated the applicability of DL models to accurately detect helmet regulation violators even in challenging lighting and weather conditions. We employed several data augmentation techniques in the study to ensure the training data is diverse enough to help build a robust model. The proposed model was tested on 100 test videos and produced an mAP score of \textbf{0.5267}, ranking \textbf{11th} on the AI City Track 5 public leaderboard. The use of deep learning techniques for image classification tasks, such as identifying helmet-wearing riders, has enormous potential for improving road safety. The study shows the potential of deep learning models for application in smart cities and enforcing traffic regulations and can be deployed in real-time for city-wide monitoring.
   
\end{abstract}

\section{Introduction}
\label{sec:intro}

Motorcycle-related injuries are one of the leading causes of traffic-related deaths worldwide. In 1994, it was estimated that motorcycles were 11 times more likely to be involved in fatal crashes than passenger cars, which increased to 27.5 times by 2007\cite{nunn2011death}. Severe blunt force trauma is the main cause of death in motorcycle accidents, which can cause internal and external damage to the rider's body. This kind of trauma typically results in injuries to the head, neck, thorax, and other parts of the body's axial-skeletal system \cite{nunn2011death}. Wearing a helmet can reduce the likelihood of such injuries. A wide number of medical and non-medical studies have found that the use of helmets can play a significant role in reducing the severity of injuries and deaths from motorcycle crashes. In Taiwan, a study considering 8,795 motorist crashes showed that the number of injuries was reduced by 33\% after the implementation of the helmet law, along with the severity of injuries\cite{chiu2000effect}. A review of 60 U.S. studies showed that helmet law implementation increased helmet usage by 47\%, and it resulted in the reduction of death (by 29\%) and injuries (by 32\%)\cite{peng2017universal}. Another evidence-based review of 197 studies from many countries worldwide (US, Thailand, Indonesia, Italy, France, and Greece)  revealed that the death rate, the number of occurrences, and the severity of injuries from motorcycle crashes are reduced due to the use of helmet\cite{macleod2010evidence}. A 2001 study investigated the effect of helmet laws on death rates by controlling factors such as population density and temperature\cite{branas2001helmet}. The study found that states with helmet laws would likely have lower motorcycle-related deaths. 

Helmet laws can only reduce deaths and injuries if a mechanism exists to enforce the laws on motorists. An observational study from Florida showed that many helmet-wearing motorcyclists wore novelty helmets (affordable low-quality helmets that do not meet safety requirements, laws, or standards)\cite{turnerhelmet}. This points out that riders need to be monitored to make sure that they are not only wearing helmets but also wearing the right kind. Based on empirical evidence, studies from Vietnam and China concluded that the proper enforcement of legislation is necessary to ensure that motorcyclists wear helmets and they do so in a proper manner\cite{hung2006prevalence,xuequn2011prevalence}. Since enforcement of helmet laws is a crucial component of ensuring safety, this calls for developing real-time helmet usage monitoring systems among motorcyclists. 

Different methods have been tested in recent years to facilitate the monitoring of helmet use. In 2019 a study conducted in Bangkok analyzed street imagery data from 462 unique motorists. By posting Human Intelligence Tasks on Amazon Mechanical Turk, they identified the motorists from the images and detected their helmet use\cite{merali2020using,Aboah23AIC23}. The study suggested that future researchers should use machine learning to automate the process. To automate the process of helmet use, a 2013 study proposed a hybrid descriptor for features extraction consisting of Local Binary Pattern, Histograms of Oriented Gradients, and the Hough Transform descriptors\cite{silva2013automatic}. A recent study used a YOLOv3-based CNN to detect helmetless motorcyclists and their number plates\cite{premmaran2022detection}. The system also aimed to automate the process of monitoring traffic rule violators. To detect helmet law violators from pre-recorded surveillance videos, another study used a YOLOv4-based CNN\cite{fredynand2021automatic}. Their system was also designed to send an email to the helmet law violators along with the penalty. There have been remarkable advances in the field, especially regarding automating a helmet-use monitoring system. However, there still is a shortfall in the number of studies that propose real-time monitoring and detection methods for detecting helmet law violations by motorists.

In this study, we developed a framework specifically for automatically detecting violations of helmet rules by motorcyclists to tackle the 2023 AI City Track 5 Challenge. The proposed methodology used an augmented annotation pipeline that pre-annotates the training dataset using an object detection model trained on the COCO dataset. These annotations are then utilized to build a helmet detection model that relies on the YOLOv5 architecture. Next, we estimate the background of each traffic video by computing the median of frames randomly sampled from a uniform distribution over twenty seconds. Our approach involves classifying motorcycles with a maximum of 3 riders on extracted backgrounds for helmet detection and violation detection. The task is to separately identify each rider on a motorcycle (i.e., driver, passenger 1, passenger 2) and determine whether they are wearing a helmet.
The 2023 AI CITY CHALLENGE provided the necessary data to train and test our proposed automatic detection system for detecting helmet violations among motorcyclists. Our proposed model was evaluated using the mean Average Precision (mAP) metric across all video frames. The mAP is a measurement that calculates the mean of average precision, which is the area under the Precision-Recall curve for all object classes as defined in the PASCAL VOC 2012 competition.

Our experimental results demonstrate that our proposed framework is effective and robust in automatically detecting helmet violations among motorcyclists.  Furthermore, our results demonstrate great potential for applicability in real-world scenarios, considering the challenges presented by road types, traffic, camera angles, lighting, and weather conditions.

\section{Related Work}


Various computer vision and image processing techniques have been used to analyze images and video sequences to detect objects, including safety helmets. These approaches can be broadly grouped into two, namely machine learning methods and deep learning methods. 

\vspace{0.1in}
\noindent\textbf{Machine Learning Approach.} \cite{wen2003safety} proposed a safety helmet detection system for ATM surveillance using a modified Hough transform to detect whether individuals in surveillance footage were wearing helmets. The proposed system uses a modified Hough transform that combines edge detection and gradient direction to identify the circular shape of safety helmets. The authors evaluated the system's performance using a dataset of ATM surveillance footage and reported high detection accuracy and low false positive rates. The main limitation of this study is that it relies exclusively on geometric properties to detect safety helmets in the image, which may not be adequate for accurate identification. Due to the similarity in shape between safety helmets and human heads, there is a possibility of confusion between the two. \cite{chiverton2012helmet} proposed a system using Histogram Oriented Gradient (HOG) features to detect motorcycles and track their movements over time. Once a motorcycle is detected, the system analyzes the corresponding rider region to determine whether a helmet is present. This is achieved using a support vector machine (SVM) classifier trained with histograms from the image data in the head region of the motorcyclists computed by the HOG descriptor\cite{aboah2023driver,aboah2021comparative,aboah2023ai}. This technique, however, does not distinguish each rider or count people on a motorcycle.  To automatically detect motorcycle riders and determine whether they are wearing safety helmets, \cite{waranusast2013machine} proposed a 4-step process that detects the presence of a motorcycle and eventually classifies each person on it. The system separates moving objects from stationary objects and extracts three features: the area of the bounding rectangle that contains the image, the aspect ratio between the width and the height of the rectangle, and the standard deviation of the hue around a rectangle at the center of the object. After all 3 features are extracted from the moving object, the K-Nearest Neighbor (KNN)  classifier is applied to these features to classify whether the object is a motorcycle or another moving object. The final step involves head extraction and classification. The primary advantage of this study was counting passengers on a motorcycle and the eventual detection of a helmet.  


\setlength{\parskip}{0pt}
\vspace{0.1in}
\noindent\textbf{Deep Learning Approach.} More recently, advanced techniques in Deep Learning to accurately detect a motorcycle's presence have been used. Most of these techniques have leveraged Convolutional Neural Networks (CNN), Region-based Convolutional Neural Networks (RCNN) \cite{jia2021real} \cite{dequito2021vision} as their backbone and have developed models that have been fine-tuned in their efficiency and accuracy in the real-time detection of motorcycle helmets.  These models include the You Only Look Once (YOLO) based network, which has evolved, EfficientDet \cite{ubaid2021automatic}, and RetinaNet, among others. The object detection algorithm YOLO9000 from \cite{redmon2017yolo9000} was used to detect the number of motorcycles in each frame of a prerecorded video clip and extract those clips with the highest number of motorcycles in them, and the RetinaNet from \cite{lin2017focal} was used for helmet use detection task. \cite{rohith2019efficient} used the Caffe model for motorcycle detection and extraction and subsequently used the Inception V3 model for helmet use classification. The proposed models showed a validation accuracy score of 86\% for motorcycle detection and 74\% for helmet use classification.   The YOLO model is yet another model that has been used in several recent studies for motorcycle detection and helmet use identification which has evolved in efficiency over time. \cite{jia2021real} introduced the improved YOLOv5\cite{aboah2020smartphone} model that was used to detect helmet use on motorcyclists automatically in real time. The method consists of two stages: motorcycle and helmet detection, and can effectively improve the precision and recall of helmet detection. The model was tested on a large-scale motorcycle helmet dataset (HFUT-MH) obtained from traffic monitoring of many cities in China, including different illumination, different perspectives, and different congestion levels. 

\setlength{\parskip}{0pt}
As discussed above, most of these studies do not have a mechanism that works in real-time to detect motorists, classify the passengers from drivers and detect the presence of a helmet. Also, these models have been trained on specific data sets in a particular geographic area and may not necessarily be applicable in different jurisdictions. This paper presents a state-of-the-art model that works in real-time to detect helmets on motorcyclists and distinguish between the passenger and driver on a motorcycle. Statistical data sampling techniques have been employed to reduce background noise in images captured and remove duplicate images before detection.

\section{Data}
\subsection{Data Overview}
\begin{figure}[h]
    \centering
    \includegraphics[width=8.3cm]{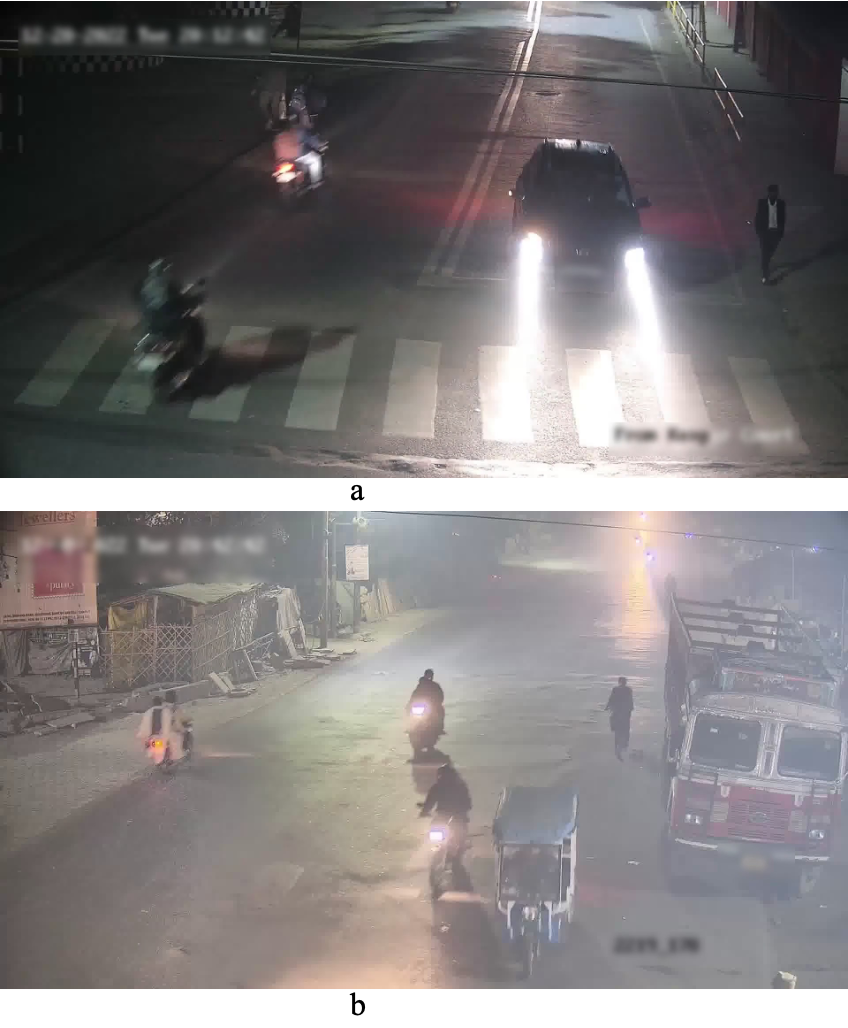}
    \caption{Visual complexities at a) night and b) foggy conditions.}
    \label{fig:com}
\end{figure}

The dataset in this competition comprises 100 videos recorded in India at a resolution of 1920x1080. The dataset poses several challenges due to the diverse visual complexities encountered under different weather conditions and times of the day, as shown in Fig.~\ref{fig:com}. Additionally, the objects of interest in the images presented additional difficulties, such as occlusion and pixelation.  Each video in the dataset was 20 seconds in length and sampled at 10 frames per second. The dataset consisted of seven classes of interest, namely (1) motorcycle, (2) helmet-wearing driver, (3) driver without a helmet, (4) first passenger wearing a helmet, (5) first passenger without a helmet, (6) second passenger wearing a helmet, and (7) second passenger without a helmet.

\subsection{Data Processing}
\begin{figure}[h]
    \centering
    \includegraphics[width=8.3cm]{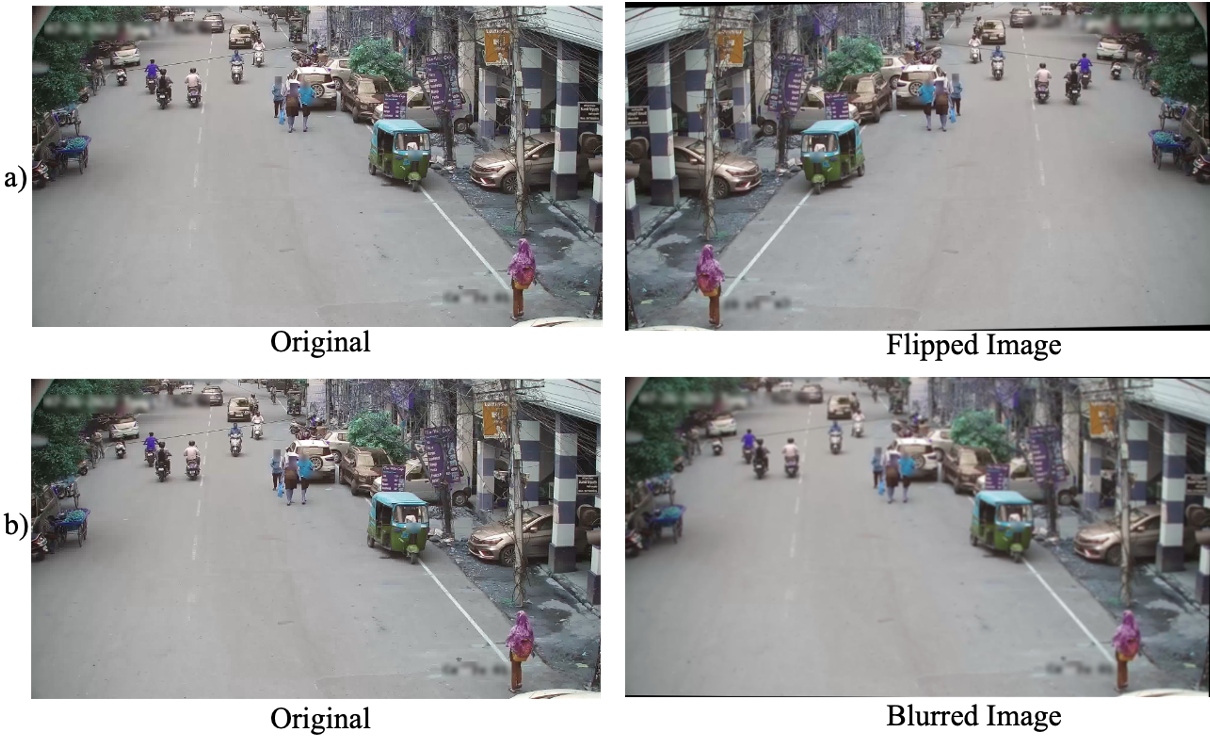}
    \caption{Augmentation Strategies a) Flipping and b) Blurring.}
    \label{fig:aug}
\end{figure}

\begin{figure}[h]
    \centering
    \includegraphics[width=8.3cm]{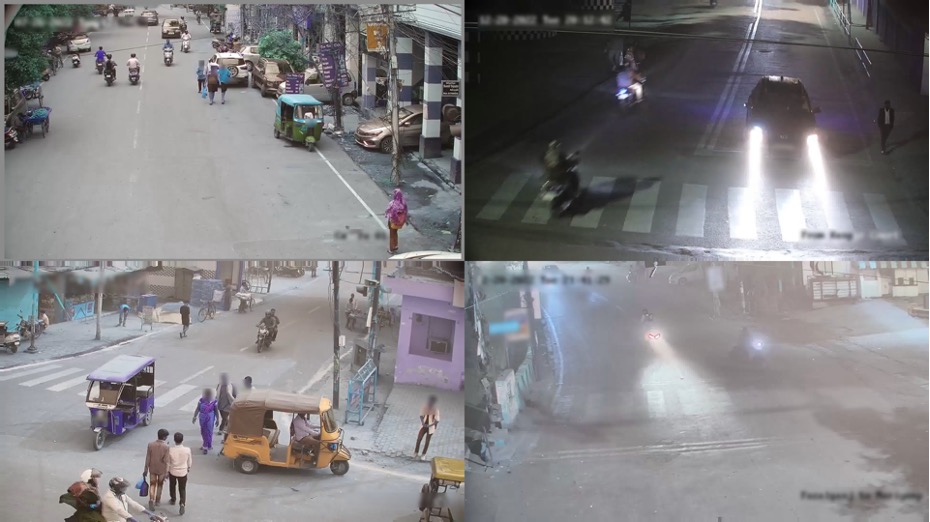}
    \caption{Sample of Mosaic Augmentation.}
    \label{fig:aug1}
\end{figure}

Several data augmentation techniques were utilized to enhance the accuracy of the detection and develop a more generalized model. These techniques comprise rotation, flipping, mosaic, and blur. The rotation process involves altering the original image's orientation at varying angles. On the other hand, flipping involves creating a mirror image of the original image along either the horizontal or vertical axis, as illustrated in Fig.~\ref{fig:aug}a. Applying the blur technique decreases the sharpness of the image by implementing a filter. The mosaic technique was specifically employed to enhance the quality of the data. This strategy necessitates resizing four distinct images and combining them to produce a mosaic image, as shown in Fig.~\ref{fig:aug1}. From this, a random segment of the mosaic image is extracted and utilized as the final augmented image. The primary benefit of this technique is that it enhances the visual complexity of the images, providing a more realistic and challenging environment for the model to recognize. By utilizing these different techniques for data augmentation, the model can tackle a broader range of images, resulting in improved accuracy in detecting the classes of interest within the dataset. 

\section{Experiment}
The present study involves solving a problem of object detection and classification. Before beginning the experiment, the training dataset was thoroughly examined to identify any potential issues. Upon inspection, we discovered several problems with the dataset, including misclassifications, false detections, and missed annotations. To resolve these issues, we reannotated selected images using the Computer Vision Annotation Tool (CVAT).

\vspace{0.1in}
\noindent\textbf{YOLOv5.} The YOLOv5\cite{aboah2021vision,boah2021mobile,shoman2022region} (You Only Look Once version 5) architecture is a deep learning-based object detection system that is designed to detect objects in real-time. The architecture follows a similar approach as its predecessors, YOLOv3 and YOLOv4, but introduces several improvements that make it more efficient and accurate. The architecture is built on a neural network that consists of a backbone and a detection head, as shown Fig.~\ref{sec:intro}. The backbone is a feature extractor that processes the input image and generates a feature map. The detection head then takes the feature map as input and predicts the bounding boxes, objectness score, and class probabilities for each object in the image. The backbone of YOLOv5 is a modified version of the EfficientNet architecture known for its efficiency and accuracy. The YOLOv5 backbone, CSPNet, consists of a series of convolutional layers and a bottleneck block that reduces the number of channels. This is followed by a cross-stage partial connection (CSP) block, which allows information to flow across the network more efficiently by splitting the feature map and processing it in parallel. The detection head of YOLOv5 comprises three convolutional layers, followed by a global average pooling layer and a fully connected layer. The output of the fully connected layer is then used to predict the bounding boxes, objectness score, and class probabilities for each object in the image. One of the key improvements introduced in YOLOv5 is the use of a novel training methodology called AutoML, which automatically selects the best hyperparameters for the network. This significantly reduces the training time and improves the accuracy of the model.

\begin{figure}[h]
    \centering
    \includegraphics[width=8.3cm]{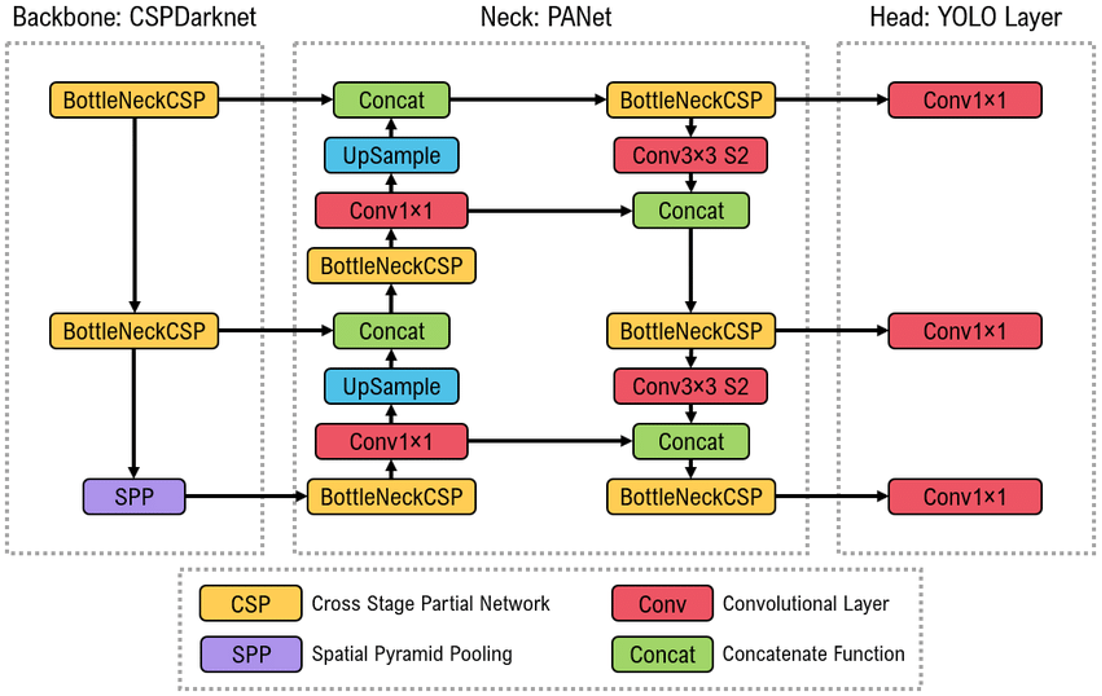}
    \caption{YOLOv5 Architecture.}
    \label{fig:yolo}
\end{figure}

\vspace{0.1in}
\noindent\textbf{Model Training.} The training dataset utilized in this experiment was divided into a ratio of 4:1, resulting in 3,482 samples for training and 802 images for validation. The training process employed a framework involving five distinct models, each with different hyperparameters. A visual representation of the training framework used in this experiment is illustrated in Figure 4. This approach was adopted to enhance the results obtained by creating five models, which could be combined using an Ensemble Deep Learning testing technique. 

\begin{table}[h]
\centering
\caption{Training Hyperparaters of All Five Models}
\label{tab:my-table}
\resizebox{\columnwidth}{!}{%
\begin{tabular}{@{}llllll@{}}
\toprule
\textbf{Hyperparameter}        & \textbf{Model 1} & \textbf{Model 2} & \textbf{Model 3} & \textbf{Model 4} & \textbf{Model 5} \\ \midrule
Initial learning rate & 0.001   & 0.01    & 0.01    & 0.001   & 0.01    \\
Image size            & 640     & 832     & 832     & 832     & 640     \\
Optimizer             & SGD     & Adam    & Adam    & SGD     & Adam    \\
Epochs                & 500     & 300     & 400     & 500     & 400     \\
Momentum              & 0.947   & 0.995   & 0.9     & 0.955   & 0.97    \\
Weight decay          & 0.0005  & 0.0005  & 0.0005  & 0.0005  & 0.0005  \\
Warmup epochs         & 3       & 5       & 4       & 5       & 7       \\
IoU                   & 0.7     & 0.9     & 0.8     & 0.9     & 0.9     \\
\bottomrule
\end{tabular}}
\end{table}

\vspace{0.1in}
\noindent\textbf{Model Evaluation.} The performance of our model was evaluated using the mean Average Precision (mAP) metric, which is derived by averaging the Average Precision (AP) scores for every frame in the test videos. The leaderboard ranked the submissions based on their mAP scores, which were calculated using Equation ~\ref{eq:map}.
\begin{equation}
\label{eq:map}
    m A P=\frac{1}{N} \sum_{i=1}^N A P_i
\end{equation}

where N is the number of queries.

\begin{figure}[h]
    \centering
    \includegraphics[width=8.3cm]{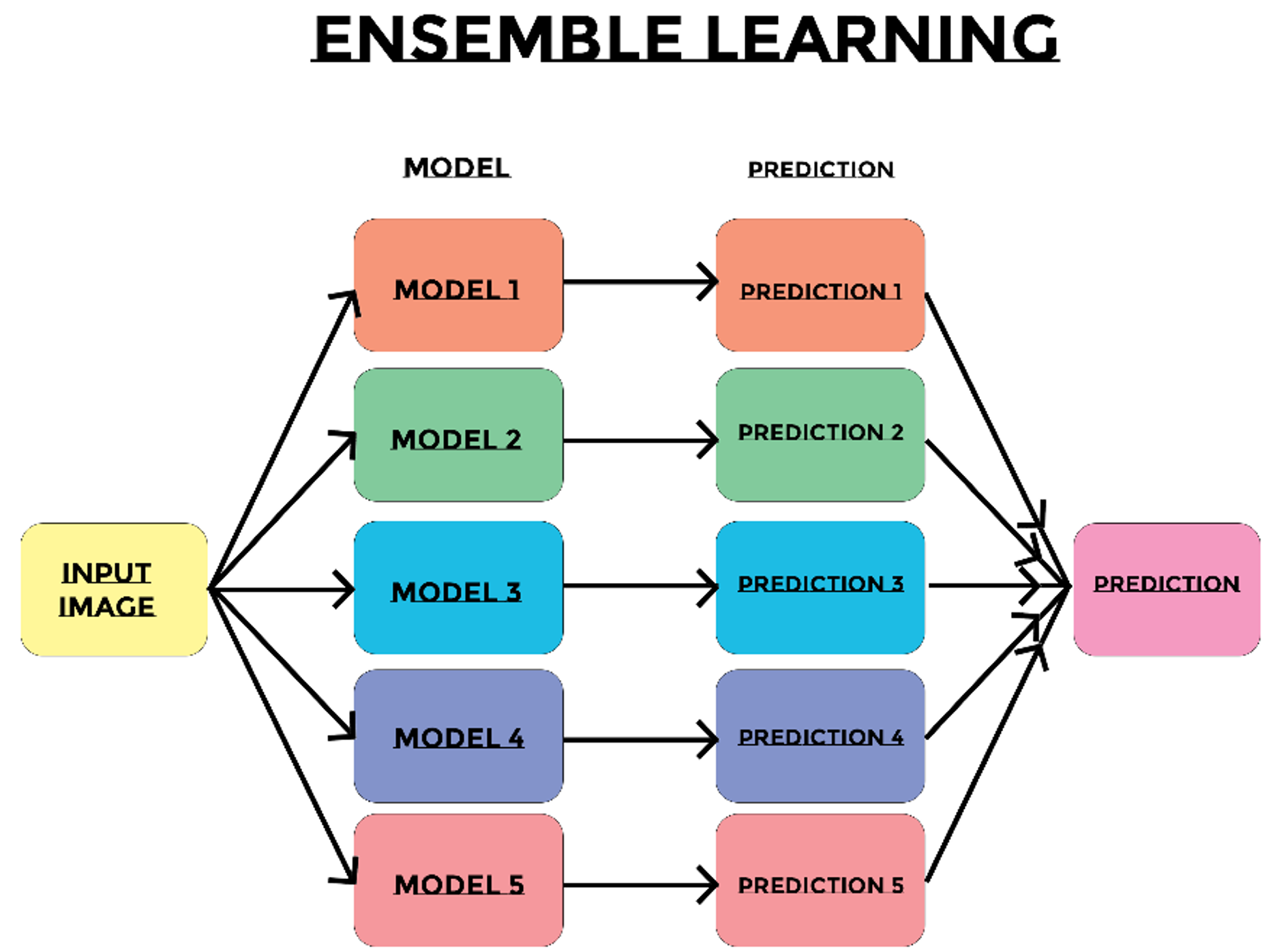}
    \caption{Testing Framework.}
    \label{fig:tt}
\end{figure}

\vspace{0.1in}
\noindent\textbf{Testing Framework.} Fig.~\ref{fig:tt} illustrates the testing framework used in our study. To perform the testing, we utilized an Ensemble Deep Learning Approach that incorporated the five YOLOv5 models we trained. This approach entailed inputting each test image into multiple pre-trained YOLOv5 models that had shown high performance. The detections from the different models were averaged to obtain class predictions at the output. 

\section{Results and Discussion}
Track 5 of the 2023 AI City Challenge provided 100 videos each recorded at 10fps at a resolution of 1920×1080 for training and testing. Each test video, just like the training videos, is of 20 seconds in duration. The objective was to detect and classify objects into the seven classes mentioned earlier.  

A submission file containing the test results in a text format follows the format: video\_id, frame, bb\_left, bb\_top, bb\_width, bb\_height, class, and confidence. The video\_id is the video numeric identifier, starting with 1, it represents the position of the video in the list of all the videos sorted in alphanumeric order. The frame refers to the frame count for the current frame in the current video, also starting with 1 sorted in alphanumeric order. The bb\_left and bb\_top are the x-coordinate and y-axis respectively of the top point of the predicted bounding box. Likewise, the bb\_width and bb\_height are the width and height respectively of the predicted bounding box. Finally, the class and confidence (a value ranging from 0 to 1) represent the predicted class and the model’s confidence score of the bounding box. Table 3 below shows a sample of the submission file format.  

\begin{figure*}[ht!]
    \centering
    \includegraphics[width=17cm]{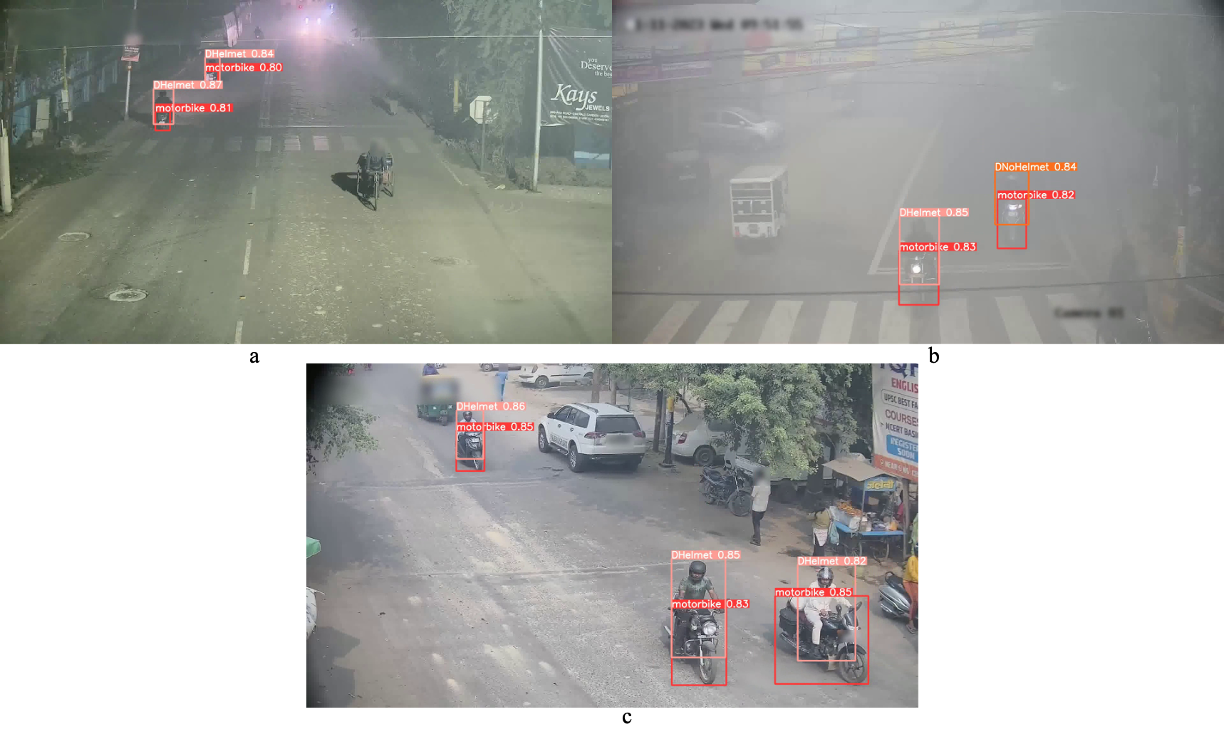}
    \caption{Model detection under varying weather conditions and time of day. a) Night, b) Foggy, and c) Daytime.}
    \label{fig:pred_img}
\end{figure*}

The evaluation for this track is the mean Average Precision (mAP) across all classes. Our model achieved an mAP of \textbf{0.5267} on 100\% of the testing dataset. This score was \textbf{11th placed} on the competition public leaderboard. Fig.~\ref{fig:pred_img} shows some detections from the model. The mAP compares the testing data bounding box information to the ground truth bounding box information and returns a score.

\section{Conclusion}
In this paper, we presented the development and evaluation of an ensemble deep-learning model using YOLOv5 for detecting and classifying motorbike passengers based on their helmet use. We employed an augmented annotation pipeline that pre-annotates training data using an object detection model trained on the COCO dataset. An ensemble deep learning of five distinct models with varying hyperparameters was further used to train the model after the application of several data augmentation techniques on the training set.

The result shows a high mAP score of 0.526 on the test data, correctly labeling the majority of the classes regardless of the lightning or weather conditions of the video. This model can also be efficiently deployed in real-time to monitor the use of helmets within city traffic. The deployment of such a tool utilizing the model for monitoring will assist law enforcement agencies and road safety authorities in enforcing helmet-wearing regulations and in turn improve the overall safety of highways by reducing cases of severe crashes in motorbike incidents.

Overall, the success of our model demonstrates the potential of deep learning techniques for addressing important real-world problems. Further improvement can be obtained by training on more datasets collected from different locations across the world. We hope that our work will inspire further research in this area, and lead to the development of even more accurate and effective models for improving safety for motorbike riders, passengers, and others on the highway. 

{\small
\bibliographystyle{unsrt}
\bibliography{egbib}

\begin{thebibliography}{10}

\bibitem{nunn2011death}
Samuel Nunn.
\newblock Death by motorcycle: background, behavioral, and situational
  correlates of fatal motorcycle collisions.
\newblock {\em Journal of forensic sciences}, 56(2):429--437, 2011.

\bibitem{chiu2000effect}
Wen-Ta Chiu, Chia-Ying Kuo, Ching-Chang Hung, and Marcelo Chen.
\newblock The effect of the taiwan motorcycle helmet use law on head injuries.
\newblock {\em American journal of public health}, 90(5):793, 2000.

\bibitem{peng2017universal}
Yinan Peng, Namita Vaidya, Ramona Finnie, Jeffrey Reynolds, Cristian Dumitru,
  Gibril Njie, Randy Elder, Rebecca Ivers, Chika Sakashita, Ruth~A Shults,
  et~al.
\newblock Universal motorcycle helmet laws to reduce injuries: a community
  guide systematic review.
\newblock {\em American journal of preventive medicine}, 52(6):820--832, 2017.

\bibitem{macleod2010evidence}
Jana~BA MacLeod, J~Christopher DiGiacomo, and Glen Tinkoff.
\newblock An evidence-based review: helmet efficacy to reduce head injury and
  mortality in motorcycle crashes: East practice management guidelines.
\newblock {\em Journal of Trauma and Acute Care Surgery}, 69(5):1101--1111,
  2010.

\bibitem{branas2001helmet}
Charles~C Branas and M~Margaret Knudson.
\newblock Helmet laws and motorcycle rider death rates.
\newblock {\em Accident Analysis \& Prevention}, 33(5):641--648, 2001.

\bibitem{turnerhelmet}
Patricia~A Turner and Christopher~A Hagelin.
\newblock Helmet use by motorcyclists: Florida observational survey results.

\bibitem{hung2006prevalence}
Dang~Viet Hung, Mark~R Stevenson, and Rebecca~Q Ivers.
\newblock Prevalence of helmet use among motorcycle riders in vietnam.
\newblock {\em Injury prevention}, 12(6):409--413, 2006.

\bibitem{xuequn2011prevalence}
Yu~Xuequn, Liang Ke, Rebecca Ivers, Wei Du, and Teresa Senserrick.
\newblock Prevalence rates of helmet use among motorcycle riders in a developed
  region in china.
\newblock {\em Accident Analysis \& Prevention}, 43(1):214--219, 2011.

\bibitem{merali2020using}
Hasan~S Merali, Li-Yi Lin, Qingfeng Li, and Kavi Bhalla.
\newblock Using street imagery and crowdsourcing internet marketplaces to
  measure motorcycle helmet use in bangkok, thailand.
\newblock {\em Injury prevention}, 26(2):103--108, 2020.

\bibitem{Aboah23AIC23}
Armstrong Aboah, Bin Wang, Bagci Ulas, and Yaw Adu-Gyamfi.
\newblock Real-time multi-class helmet violation detection using few-shot data
  sampling technique and yolov8.
\newblock In {\em The IEEE Conference on Computer Vision and Pattern
  Recognition (CVPR) Workshops}, June 2023.

\bibitem{silva2013automatic}
Romuere Silva, Kelson Aires, Thiago Santos, Kalyf Abdala, Rodrigo Veras, and
  Andr{\'e} Soares.
\newblock Automatic detection of motorcyclists without helmet.
\newblock In {\em 2013 XXXIX Latin American computing conference (CLEI)}, pages
  1--7. IEEE, 2013.

\bibitem{premmaran2022detection}
G~Premmaran and P~Sathishkumar.
\newblock Detection of helmetless riders and automatic number plate recognition
  using machine learning.
\newblock In {\em 2022 International Conference on Applied Artificial
  Intelligence and Computing (ICAAIC)}, pages 339--345. IEEE, 2022.

\bibitem{fredynand2021automatic}
Colin Fredynand, Irene~Elizabeth John, Milan Koshy, and Sharon~Binu George.
\newblock Automatic detection of helmet violation.
\newblock 2021.

\bibitem{wen2003safety}
Che-Yen Wen, Shih-Hsuan Chiu, Jiun-Jian Liaw, and Chuan-Pin Lu.
\newblock The safety helmet detection for atm's surveillance system via the
  modified hough transform.
\newblock In {\em IEEE 37th Annual 2003 International Carnahan Conference
  onSecurity Technology, 2003. Proceedings.}, number~1, pages 364--369. IEEE,
  2003.

\bibitem{chiverton2012helmet}
John Chiverton.
\newblock Helmet presence classification with motorcycle detection and
  tracking.
\newblock {\em IET Intelligent Transport Systems}, 6(3):259--269, 2012.

\bibitem{aboah2023driver}
Armstrong Aboah, Yaw Adu-Gyamfi, Senem~Velipasalar Gursoy, Jennifer Merickel,
  Matt Rizzo, and Anuj Sharma.
\newblock Driver maneuver detection and analysis using time series segmentation
  and classification.
\newblock {\em Journal of Transportation Engineering, Part A: Systems},
  149(3):04022157, 2023.

\bibitem{aboah2021comparative}
Armstrong Aboah and Elizabeth Arthur.
\newblock Comparative analysis of machine learning models for predicting travel
  time.
\newblock {\em arXiv preprint arXiv:2111.08226}, 2021.

\bibitem{aboah2023ai}
Armstrong Aboah, Abdul~Rashid Mussah, and Yaw Adu-Gyamfi.
\newblock Ai-based framework for understanding car following behaviors of
  drivers in a naturalistic driving environment.
\newblock {\em arXiv preprint arXiv:2301.09315}, 2023.

\bibitem{waranusast2013machine}
Rattapoom Waranusast, Nannaphat Bundon, Vasan Timtong, Chainarong Tangnoi, and
  Pattanawadee Pattanathaburt.
\newblock Machine vision techniques for motorcycle safety helmet detection.
\newblock In {\em 2013 28th International conference on image and vision
  computing New Zealand (IVCNZ 2013)}, pages 35--40. IEEE, 2013.

\bibitem{jia2021real}
Wei Jia, Shiquan Xu, Zhen Liang, Yang Zhao, Hai Min, Shujie Li, and Ye~Yu.
\newblock Real-time automatic helmet detection of motorcyclists in urban
  traffic using improved yolov5 detector.
\newblock {\em IET Image Processing}, 15(14):3623--3637, 2021.

\bibitem{dequito2021vision}
CJM Dequito, IJL Dichaves, RJG Juan, MYKT Minaga, JP~Ilao, II~MO Cordel, and
  NPA Del~Gallego.
\newblock Vision-based bicycle and motorcycle detection using a yolo-based
  network.
\newblock In {\em Journal of Physics: Conference Series}, volume 1922, page
  012003. IOP Publishing, 2021.

\bibitem{ubaid2021automatic}
Muhammad~Talha Ubaid, Amara Kiran, Muhammad~Tayyab Raja, Umme~Aliza Asim, Abdou
  Darboe, and Muhammad~Asad Arshed.
\newblock Automatic helmet detection using efficientdet.
\newblock In {\em 2021 International Conference on Innovative Computing
  (ICIC)}, pages 1--9. IEEE, 2021.

\bibitem{redmon2017yolo9000}
Joseph Redmon and Ali Farhadi.
\newblock Yolo9000: better, faster, stronger.
\newblock In {\em Proceedings of the IEEE conference on computer vision and
  pattern recognition}, pages 7263--7271, 2017.

\bibitem{lin2017focal}
Tsung-Yi Lin, Priya Goyal, Ross Girshick, Kaiming He, and Piotr Doll{\'a}r.
\newblock Focal loss for dense object detection.
\newblock In {\em Proceedings of the IEEE international conference on computer
  vision}, pages 2980--2988, 2017.

\bibitem{rohith2019efficient}
CA~Rohith, Shilpa~A Nair, Parvathi~Sanil Nair, Sneha Alphonsa, and
  Nithin~Prince John.
\newblock An efficient helmet detection for mvd using deep learning.
\newblock In {\em 2019 3rd International Conference on Trends in Electronics
  and Informatics (ICOEI)}, pages 282--286. IEEE, 2019.

\bibitem{aboah2020smartphone}
Armstrong Aboah and Yaw Adu-Gyamfi.
\newblock Smartphone-based pavement roughness estimation using deep learning
  with entity embedding.
\newblock {\em Advances in Data Science and Adaptive Analysis},
  12(03n04):2050007, 2020.

\bibitem{aboah2021vision}
Armstrong Aboah.
\newblock A vision-based system for traffic anomaly detection using deep
  learning and decision trees.
\newblock In {\em Proceedings of the IEEE/CVF Conference on Computer Vision and
  Pattern Recognition}, pages 4207--4212, 2021.

\bibitem{boah2021mobile}
Armstrong Aboah, Michael Boeding, and Yaw Adu-Gyamfi.
\newblock Mobile sensing for multipurpose applications in transportation.
\newblock {\em arXiv preprint arXiv:2106.10733}, 2021.

\bibitem{shoman2022region}
Maged Shoman, Armstrong Aboah, Alex Morehead, Ye~Duan, Abdulateef Daud, and Yaw
  Adu-Gyamfi.
\newblock A region-based deep learning approach to automated retail checkout.
\newblock In {\em Proceedings of the IEEE/CVF Conference on Computer Vision and
  Pattern Recognition}, pages 3210--3215, 2022.

\end{thebibliography}
}

\end{document}